\documentclass{article}
\usepackage{iclr2027_conference,times}
\iclrfinalcopy  
\usepackage[T1]{fontenc}
\usepackage{textcomp}

\usepackage{amsmath,amssymb}
\usepackage{graphicx}
\usepackage{float}
\usepackage{placeins}
\usepackage{booktabs}
\usepackage{multirow}
\usepackage{array}
\usepackage{xcolor}
\usepackage{colortbl}
\usepackage{enumitem}
\setlist[itemize]{itemsep=2pt,topsep=3pt,parsep=0pt}
\usepackage{fancyvrb}
\usepackage{wrapfig}

\usepackage[most]{tcolorbox}
\usepackage{soul}
\usepackage{hyperref}
\usepackage{fontawesome5}
\usepackage{url}

\definecolor{accent}{RGB}{0,108,102}
\definecolor{keyfill}{HTML}{EEF2F7}   
\definecolor{keytext}{HTML}{2B4A6F}   
\definecolor{promptfill}{HTML}{FFFFF8}  
\hypersetup{colorlinks=true, citecolor=accent, urlcolor=accent, linkcolor=black}

\definecolor{navy}{RGB}{28,52,94}
\definecolor{navytint}{RGB}{236,241,248}
\definecolor{teal}{RGB}{30,128,126}
\definecolor{coral}{RGB}{211,91,73}
\definecolor{gold}{RGB}{194,139,44}
\newtcolorbox{promptbox}[1][Prompt]{enhanced, breakable,
  colback=white, colframe=navy!55, boxrule=0.6pt,
  arc=2pt, outer arc=2pt,
  left=8pt, right=8pt, top=6pt, bottom=6pt,
  fonttitle=\bfseries, coltitle=navy, colbacktitle=navytint,
  toptitle=4pt, bottomtitle=4pt, titlerule=0.4pt,
  title={#1},
}

\usepackage{xspace}

\definecolor{phPlanning}{HTML}{2F77C7}
\definecolor{phExec}{HTML}{2F9E6F}
\definecolor{phRestate}{HTML}{C68A12}
\definecolor{phFact}{HTML}{8257C9}
\definecolor{phReflect}{HTML}{CF5D86}
\definecolor{phConclude}{HTML}{1E97A6}

\definecolor{phPlanningBg}{HTML}{D8E7FA}
\definecolor{phExecBg}{HTML}{D5F0E2}
\definecolor{phRestateBg}{HTML}{F9EAC6}
\definecolor{phFactBg}{HTML}{E5D9F9}
\definecolor{phReflectBg}{HTML}{FAD8E6}
\definecolor{phConcludeBg}{HTML}{CDEDF1}

\newcommand{\spanP}[1]{{\sethlcolor{phPlanningBg}\hl{#1}}}
\newcommand{\spanE}[1]{{\sethlcolor{phExecBg}\hl{#1}}}
\newcommand{\spanS}[1]{{\sethlcolor{phRestateBg}\hl{#1}}}
\newcommand{\spanF}[1]{{\sethlcolor{phFactBg}\hl{#1}}}
\newcommand{\spanV}[1]{{\sethlcolor{phReflectBg}\hl{#1}}}
\newcommand{\spanC}[1]{{\sethlcolor{phConcludeBg}\hl{#1}}}

\newcommand{\pw}[1]{\textsuperscript{#1}}
\newsavebox{\phgebox}\sbox{\phgebox}{$\ge$}
\newcommand{\GE}{\usebox{\phgebox}}
\newsavebox{\phtimesbox}\sbox{\phtimesbox}{$\times$}
\newcommand{\TIMES}{\usebox{\phtimesbox}}
\soulregister{\pw}{1}
\soulregister{\GE}{0}
\soulregister{\TIMES}{0}

\newcommand{\nmPlanning}{\textcolor{phPlanning!75!black}{\emph{planning}}}
\newcommand{\nmExec}{\textcolor{phExec!70!black}{\emph{execution}}}
\newcommand{\nmRestate}{\textcolor{phRestate!75!black}{\emph{restatement}}}
\newcommand{\nmFact}{\textcolor{phFact!75!black}{\emph{fact}}}
\newcommand{\nmReflect}{\textcolor{phReflect!80!black}{\emph{reflection}}}
\newcommand{\nmConclude}{\textcolor{phConclude!75!black}{\emph{conclusion}}}

\newcommand{\elide}[1]{\par\vspace{3pt}\centerline{\footnotesize\color{black!45}$\cdots$\ #1\ $\cdots$}\vspace{2pt}\par}

\tcbset{phasebox/.style={
  enhanced, colback=white, colframe=navy!55, boxrule=0.7pt,
  arc=2.5pt, outer arc=2.5pt, left=9pt, right=9pt, top=6pt, bottom=7pt,
  fonttitle=\bfseries\sffamily, coltitle=navy, colbacktitle=navytint,
  toptitle=4pt, bottomtitle=4pt, titlerule=0.4pt}}

\title{Routes of Reasoning: Reading and Reshaping Test-Time Compute in MoE Models}

\makeatletter
\newcommand{\codefootnote}[1]{\begingroup\renewcommand{\thefootnote}{}%
  \renewcommand{\@makefntext}[1]{\noindent##1}\footnotetext{#1}\endgroup}
\makeatother
\newcommand{\codeurl}{https://github.com/swang-ai/reasoning-moe}
\author{Shengjie Wang,\ \ Mohsen Fayyaz,\ \ Nanyun Peng\\[2pt]
{\normalfont University of California, Los Angeles}}

\begin{document}
\maketitle
\lhead{Preprint}
\codefootnote{\href{\codeurl}{\textcolor{accent}{\faGithub\ Code}}}

\begin{abstract}
  Reasoning models allocate test-time compute across distinct behaviors such as planning, deriving,
  and verifying. However, models frequently ``overthink,'' continuing to verify and re-derive long
  after finding a correct answer. This wastes expensive compute and can even degrade performance.
  Existing interventions, such as token budgets or steering, primarily regulate reasoning duration
  rather than targeting specific behaviors.
  In this work, we demonstrate that Mixture-of-Experts (MoE) routers provide a natural, training-free
  interface for behavioral control. We first split reasoning traces into sentence-level
  segments and label each with one of six reasoning behaviors (\nmPlanning{}, \nmExec{}, \nmRestate{}, \nmFact{},
  \nmReflect{}, and \nmConclude{}), so that every token carries the behavior of its segment. Linear probes
  on router logits predict this label nearly as well as probes on the residual stream, the model's full
  hidden state and the standard target for probing, while using about half as many features. These
  signals generalize: experts associated with a reasoning behavior on one dataset predict the same
  behavior on the others, across six datasets spanning five domains (math, code, science, knowledge,
  and commonsense). Building on this, we selectively edit expert router
  scores during inference to curb overthinking. Across four models and four benchmarks, this
  intervention reduces the number of generated tokens by a median of 30.4\% (and up to 63.3\%).
  Crucially, it encourages the model to stop once the problem is solved, rather than skipping
  essential reasoning steps. Accuracy never drops by more than 0.5 percentage points, and frequently
  improves.
  Ultimately, MoE routing offers both an interpretable lens into reasoning states and a lightweight
  mechanism for optimizing test-time compute.
\end{abstract}

\section{Introduction}
\label{sec:intro}

Reasoning models use additional computation at test time to solve difficult problems
\citep{guo2025deepseekr1,team2025qwen3,openai2025gptoss}. Within a single response, that computation
can serve different purposes: planning an approach, carrying out a derivation, recalling a fact, or
checking an earlier result. How computation is allocated therefore matters as much as how much is
spent. Models can also \emph{overthink}: after reaching a correct answer, they continue to verify
and re-derive it, spending tokens without advancing the solution
\citep{chen2024overthinking,sui2025stop}. Addressing this waste requires more than measuring trace
length. We need to relate what the model is doing to the computational components it uses.

In this work, we demonstrate that sparse Mixture-of-Experts (MoE) models offer a natural internal view of this allocation. At each
layer, a router selects a small subset of experts for every token. Prior work has found strong token
dependence in routing \citep{xue2024openmoe}, semantic structure in router decisions
\citep{olson2025semantic}, and expert-mediated behavior control \citep{dahlke2025mote}. We ask a
finer-grained question: does routing distinguish the \emph{functional role} of a reasoning span,
such as planning or reflection, as that role changes within a problem?

A positive answer alone would not establish a useful control interface. Router states might reflect
lexical cues or dataset identity rather than reasoning behavior; similarly, an expert set associated
with a behavior might not remain associated outside the data on which it was observed. We therefore
study a sequence of increasingly demanding questions: can routing read out reasoning behavior, do
behavior--expert associations persist across domains, and can they guide useful test-time
interventions?

\begin{figure}[!t]
\centering
\definecolor{figbg}{HTML}{FAF7F0}
\begin{tcolorbox}[colback=figbg, colframe=figbg, boxrule=0pt, arc=7pt,
  left=5pt, right=5pt, top=5pt, bottom=5pt]
\includegraphics[width=\linewidth]{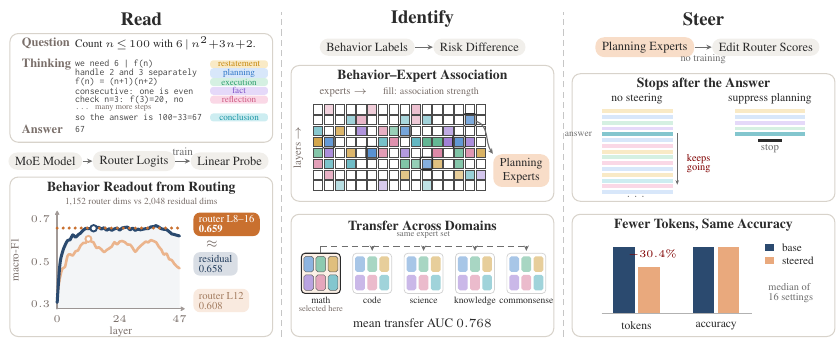}
\end{tcolorbox}
\vspace{-10pt}
\caption{\textbf{Overview.} \emph{Read}: router logits reveal the behavior of each reasoning step
(Section~\ref{sec:analysis}). \emph{Identify}: behavior-associated experts recur across domains
(Section~\ref{sec:identify}). \emph{Steer}: editing behavior-associated experts reduces generated
tokens while largely preserving accuracy (Section~\ref{sec:intervention}).}
\label{fig:teaser-ai}
\end{figure}

\noindent\textbf{First,} we annotate reasoning traces with six behaviors and test whether router
states expose the behavior of the current reasoning step. In Qwen3-30B-A3B, a compact linear router
readout approaches a probe on the substantially wider residual stream. This establishes routing as a
practical observation point for reasoning behavior (Section~\ref{sec:analysis}).

\noindent\textbf{Next,} we ask whether these associations are tied to one dataset or reflect a more
stable organization of the model. Behavior-associated expert profiles recur across six datasets
spanning mathematics, code, science, knowledge, and commonsense; expert sets selected on one source
also predict behavior on held-out domains (Section~\ref{sec:identify}).

\noindent\textbf{Finally,} we use development-selected behavior-associated expert sets to steer
routing at inference time. Across four MoE models and four benchmarks, the selected interventions
curb overthinking, reducing generated tokens by a median of 30.4\% (up to 63.3\%) while accuracy
drops by at most 0.5 points, and outperform equally sized random edits and a brevity prompt
(Section~\ref{sec:intervention}). We further analyze why suppressing
planning-associated experts is effective and where in a trace the saving comes from
(Section~\ref{sec:why-planning}).

\section{Related Work}
\label{sec:related}

\paragraph{Test-time reasoning and efficiency.}
Reasoning models use extended generations to support multi-step problem solving
\citep{guo2025deepseekr1,team2025qwen3}, but trace length alone does not determine reasoning quality:
both redundant computation and premature switching between solution paths can hurt efficiency
\citep{chen2024overthinking,wang2025underthinking,sui2025stop}. Existing approaches control token
budgets through prompting and decoding \citep{han2024tokenbudget,muennighoff2025s1}, learn
length-constrained policies \citep{aggarwal2025l1}, or edit weights to mitigate overly short reasoning
\citep{sun2025thinkedit}. Closely related, SEAL distinguishes execution, reflection, and transition
thoughts and calibrates reasoning through activation steering \citep{chen2025seal}. Our work shares
this interest in the composition of reasoning, but studies MoE routing as the interface for reading
behavior and selecting intervention targets.

\paragraph{Interpreting expert specialization.}
Analyses of open MoE models document token-dependent routing and expert specialization
\citep{xue2024openmoe,muennighoff2024olmoe}. Other studies examine sensitivity to input semantics
and collaboration among experts \citep{olson2025semantic,tang2025hidden}, or show that routing is
language-specific in early and late layers but aligned across languages in the middle layers
\citep{bandarkar2026multilingual}. Complementary work uses
sparse autoencoders to study features within reasoning-model activations \citep{hazra2025hood}.
These perspectives motivate distinguishing a span's functional role from its vocabulary or topic.
We align six reasoning-behavior labels with routing states and compare router-logit probes with
residual-stream probes, then use behavior associations to identify candidate layer--expert pairs
for intervention.

\paragraph{Inference-time expert steering.}
Expert selection and weighting are established targets for inference-time control
\citep{dahlke2025mote,li2025c3po,wang2025rice,fayyaz2026steermoe}. C3PO optimizes expert mixing
using reference examples, while RICE reinforces cognitive experts associated with thinking markers
through normalized pointwise mutual information \citep{li2025c3po,wang2025rice}. SteerMoE selects
experts from activation-frequency differences between contrasting inputs and activates or deactivates
them to control behaviors including safety and faithfulness \citep{fayyaz2026steermoe}, and
promoting middle-layer experts that are frequently used in English improves multilingual performance
\citep{bandarkar2026multilingual}.
Our intervention follows this expert (de)activation interface. What differs is how the experts are
chosen, by their association with reasoning behaviors annotated in the model's own traces, and what
we measure: the effect on reasoning length and accuracy, and on where in a trace the saving arises.

\section{Background and Problem Setup}
\label{sec:data}

We define reasoning behaviors and their alignment with sparse expert routing, providing the shared
representations used for behavior readout and expert identification.

\subsection{Sparse Mixture-of-Experts Routing}
\label{sec:moe-routing}

An MoE layer replaces the dense feed-forward sublayer with $n_r$ routed experts and, in some models,
$n_s$ always-active shared experts \citep{shazeer2017moe,dai2024deepseekmoeacl}. For token representation
$h_t^{(\ell)}$, the router produces logits $z_t^{(\ell)}=W_r^{(\ell)}h_t^{(\ell)}$ and selects the
$k$ highest-probability routed experts,
\begin{equation}
  p_t^{(\ell)} = \operatorname{softmax}\!\bigl(z_t^{(\ell)}\bigr),
  \qquad
  S_t^{(\ell)} = \operatorname{Top}\text{-}K\bigl(p_t^{(\ell)},k\bigr),
  \qquad
  \bigl|S_t^{(\ell)}\bigr| = k \ll n_r,
  \label{eq:routing}
\end{equation}
The retained probabilities are renormalized into gate weights
$g_{t,e}^{(\ell)}=p_{t,e}^{(\ell)}/\sum_{j\in S_t^{(\ell)}}p_{t,j}^{(\ell)}$, and the layer output is
\begin{equation}
  \operatorname{MoE}^{(\ell)}\!\bigl(h_t^{(\ell)}\bigr)
  = \sum_{e\in S_t^{(\ell)}} g_{t,e}^{(\ell)}
      E_e^{(\ell)}\!\bigl(h_t^{(\ell)}\bigr)
    + \sum_{j=1}^{n_s} E_j^{(s,\ell)}\!\bigl(h_t^{(\ell)}\bigr).
  \label{eq:moe-output}
\end{equation}
We call the token--layer sequence of $(z_t^{(\ell)},S_t^{(\ell)},g_t^{(\ell)})$ the \emph{routing
trace}. Expert identities are layer- and model-specific. The second term of Eq.~\eqref{eq:moe-output}
holds the always-active shared experts, which the sparse router does not select, so we exclude them
from routing analysis.

\subsection{Reasoning Behaviors and Annotation}
\label{sec:taxonomy}
\label{sec:pipeline}

We characterize a reasoning span by its functional role in solving the problem. The six behaviors in
Table~\ref{tab:phases} distinguish proposing an approach, carrying out an inference, and evaluating
previous work, alongside restatement, fact recall, and conclusion.

\begin{wraptable}{r}{0.53\textwidth}
\vspace{-\intextsep}
\caption{Six reasoning behaviors.}
\label{tab:phases}
\small
\centering
\begin{tabular}{@{}l@{\hspace{6pt}}l@{}}
\toprule
Behavior & What the segment does\\
\midrule
\nmPlanning{}    & proposes how to proceed\\
\nmExec{}        & derives a new result\\
\nmRestate{}     & repeats without new inference\\
\nmFact{}        & invokes established knowledge\\
\nmReflect{}     & checks or revises earlier work\\
\nmConclude{}    & states a result or closes a subgoal\\
\bottomrule
\end{tabular}
\vspace{-\intextsep}
\end{wraptable}

These labels describe local functions, not positions in a fixed sequence. Planning can resume after
an unsuccessful derivation, and reflection can occur before an intermediate conclusion. The distinction is contextual: the same equation may
introduce a new deduction, recall a known fact, or repeat an earlier result.

For a response $x_{1:T}$, segmentation yields contiguous spans $s_m=x_{a_m:b_m}$. An LLM judge (GPT-5.5)
assigns each span a label $y_m\in\mathcal{Y}$ using the full trace as context. Labels describe the
role of the written reasoning, rather than whether its claims are correct. We align span boundaries
with the model's tokenizer and assign $y_t=y_m$ for $a_m\leq t\leq b_m$, pairing text-derived labels
with the corresponding routing states. Segmentation settings are specified with each experiment;
the annotation protocol and codebook appear in Appendix~\ref{app:annotation}. This separates the
annotation unit from the temporal window used to aggregate features for probing.

\subsection{Routing Representations}
\label{sec:routing-features}

We use two complementary views of routing. For behavior readout (Section~\ref{sec:analysis}),
continuous pre-top-$k$ logits $z_t^{(\ell)}$ retain scores for every routed expert, including those
not selected. For expert identification (Section~\ref{sec:identify}), we use post-top-$k$
selection indicators to measure participation in the executed computation:
\begin{equation}
  r_{t,e}^{(\ell)} =
  \begin{cases}
    1, & e\in S_t^{(\ell)},\\
    0, & \text{otherwise}.
  \end{cases}
  \label{eq:sparse-router-feature}
\end{equation}
We average these features within each annotated segment, retaining layer identity. Readout comparisons
select one layer on development data; expert profiles are analyzed separately at each layer.

\section{Reading Reasoning Through the Router}
\label{sec:analysis}

Can the router reveal the kind of reasoning a model is performing? The answer decides whether the rest
of this paper is possible: if routing carries the current behavior, then the experts it selects are
candidate handles on that behavior. We study the question in Qwen3-30B-A3B and establish three things.
Behavior is readable from routing at individual tokens; a few router layers carry as much of it as
the residual stream; and what the probe reads is not a shortcut through surface wording.

\subsection{Probe Setup}
\label{sec:routing-case-study}

We generate 330 traces on a mixture of GSM8K \citep{cobbe2021gsm8k}, NuminaMath
\citep{li2024numinamath}, and AIME 2023 \citep{maa_aime} problems, yielding 78,955 labeled
sentence- or clause-level segments. Problems are split into train, dev, and test sets
(Table~\ref{tab:qwen-splits}; Appendix~\ref{app:annotation}). Each token inherits its segment's behavior label. We extract its 128
pre-top-$k$ router logits at each of the 48 layers, subtract their within-token mean, and fit an
L2-regularized multinomial logistic probe. Regularization and layer are selected on dev; we report
held-out macro-F1 with problem-bootstrap 95\% confidence intervals (Appendix~\ref{app:part1-setup}).

\subsection{Reasoning Behavior Is Decodable from Routing}
\label{sec:linear-mlp}

\paragraph{Readable at a single token.}
We first predict a token's behavior from its own router state at one layer, with no pooling and no
neighboring context. The dev-selected layer, L12 (layer indices are zero-based), reaches 0.608
macro-F1 on held-out tokens (95\% CI [0.584, 0.630]), far above the 0.096 obtained by always
predicting the most frequent behavior. The 128 scores a router assigns to its experts already
indicate what the model is doing.

\begin{figure}[!htb]
\centering
\includegraphics[width=\linewidth]{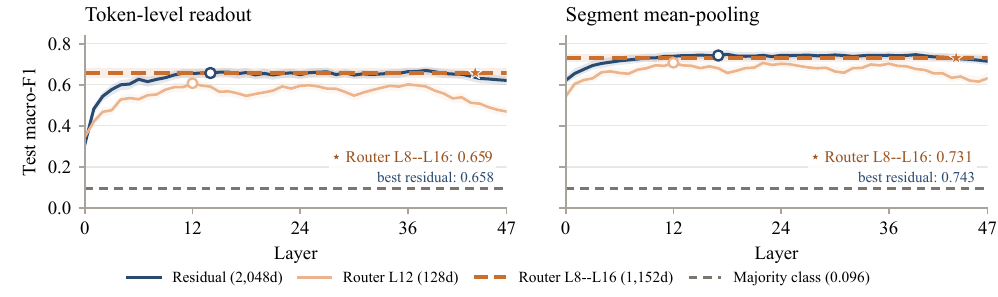}
\caption{Linear behavior readout. Curves use one layer; circles mark dev-selected layers.
The dashed line concatenates router L8--L16. Bands: 95\% problem-bootstrap CIs.}
\label{fig:router-residual-readout}
\end{figure}

\paragraph{A few router layers match the residual stream.}
How much does this compact state leave out? We repeat the scan on 2,048-dimensional residual states at
the same token positions, with the same split, preprocessing, and tuning budget. The best residual
layer, L14, reaches 0.658, five points above one router layer (Figure~\ref{fig:router-residual-readout}a).
Concatenating the router logits of a prespecified nine-layer window around L12 (L8--L16) closes the
gap: the 1,152-dimensional readout reaches 0.659, a paired gain of 5.1 points over one router layer
(95\% CI [4.1, 6.1]) and within 0.1 points of the residual stream ([$-0.8$, 1.0]) at 56\% of its width.
Parity therefore requires combining layers; a single router layer carries most, but not all, of the
signal. The match is not an artifact of drawing on more layers: residual states from the same nine
layers, compressed to the same 1,152 dimensions, reach 0.657, and the result is insensitive to the
window width (Appendix~\ref{app:router-depth}).

\subsection{Not a Shortcut Through Surface Wording}
\label{sec:analysis-controls}

A probe could score well by keying on phrases that announce a behavior, such as \emph{``Wait''} or
\emph{``Let me''}. Two checks argue against this. First, routing beats readouts built from the text
itself: in a segment-level comparison with the same neighboring context, router features reach 0.709,
above input embeddings (0.624), lexical TF-IDF (0.566), and trace position (0.124). Second, the
signal does not live in a segment's opening words. On segments with at least six tokens, pooling the
router states after dropping the first five tokens still gives 0.677, above the 0.626 obtained from
those five tokens alone and close to the 0.693 of the full segment.

\paragraph{Reading a whole segment.}
Pooling router states within a segment turns the token readout into a segment readout. Mean pooling at
L12 reaches 0.708, and pooling the nine-layer window reaches 0.731, close to the 0.743 of the best
residual layer (Figure~\ref{fig:router-residual-readout}b). Other pooling and voting schemes are
compared in Appendix~\ref{app:part1-results}.

\begin{tcolorbox}[
  colback=keyfill, colframe=keyfill, boxrule=0pt,
  arc=0pt, left=6pt, right=6pt, top=5pt, bottom=5pt
]
\textbf{\color{keytext}Takeaway.}
Which experts the router selects reveals which behavior the model is engaged in, compactly and beyond
surface wording. Section~\ref{sec:identify} asks which experts carry this signal.
\end{tcolorbox}

\section{Identifying Reasoning Experts}
\label{sec:identify}

Can behavior-associated experts be identified beyond the dataset on which they were observed?
We analyze 1,980 Qwen3-30B-A3B traces: 330 each from the mathematical mixture above,
MATH500 \citep{hendrycks2021math}, LiveCodeBench \citep{jain2024livecodebench},
GPQA (non-Diamond) \citep{rein2024gpqa}, MMLU-Pro \citep{wang2024mmlupro}, and
CommonsenseQA \citep{talmor2019commonsenseqa}. Together these cover mathematics,
code, science, knowledge, and commonsense.

\subsection{Behavior--Expert Association}
\label{sec:expert-association}

At each layer, we rank experts by the absolute increase in selection probability for behavior $b$
relative to dataset $d$ overall. This \emph{risk difference} measures behavior-specific recruitment:
\begin{equation}
 a_{b,d,e}^{(\ell)}=p(e\text{ selected}\mid b,d,\ell)
                                      -p(e\text{ selected}\mid d,\ell).
 \label{eq:expert-specificity}
\end{equation}
This controls for differences in baseline expert usage across datasets. We compare the resulting
128-expert profiles for the same versus different behaviors, and use within-trace label permutations
as a null. Layer 12, selected by the largest same-minus-different behavior correlation gap, has
same-behavior correlation 0.729, while different-behavior
correlation is $-0.107$, indicating more consistent profiles within than across behaviors.
Top-10 sets have mean cross-dataset Jaccard 0.511, compared with a permuted-label null of 0.202.
For each of planning, reflection, and conclusion, six of the top ten experts are shared by all six
datasets. These sets carry 15\%--23\% of routing mass, connecting behavioral association to
substantial participation in the routed computation.

\subsection{Behavior-Associated Experts Transfer Across Domains}
\label{sec:expert-selection}

\begin{wrapfigure}{r}{0.46\textwidth}
\vspace{-\intextsep}
\centering
\includegraphics[width=\linewidth]{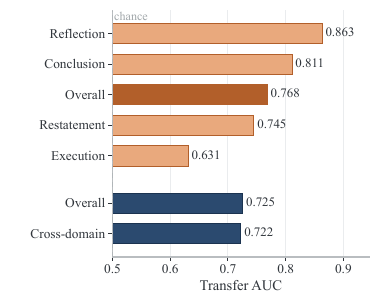}
\caption{\textbf{Expert-set transfer at layer 36.} Rose: overall AUC and four reported behaviors on
full segments. Blue-gray: the same ranking after dropping the first five tokens of each segment.
Dashed line: chance.}
\label{fig:qwen-transfer}
\vspace{-\intextsep}
\end{wrapfigure}

We select a behavior's top-10 experts on a source dataset and use their selection frequency to rank
segments on a different dataset, without using target labels to select the set. At layer 36, mean
source-to-target AUC is 0.768. Reflection and conclusion transfer particularly well (0.863 and 0.811),
followed by restatement (0.745); execution is weaker at 0.631 (Figure~\ref{fig:qwen-transfer}).

After removing the first five tokens of each eligible segment, overall AUC remains 0.7254,
with cross-domain AUC 0.7223. This suggests that opening markers alone do not explain transfer,
although lexical cues elsewhere and changes in the eligible segment population remain possible.
Separately, a linear probe using only expert-selection
fractions from four layers transfers across domains at 0.561 macro-F1, compared with 0.701 on held-out
traces within a dataset and 0.096 for a majority-class baseline. The routing signature therefore supports both expert-set ranking and supervised
behavior classification beyond the source domain.

\section{Steering Experiments}
\label{sec:intervention}

\subsection{The Steering Operator: Activation and Deactivation}
\label{sec:intervention-method}

We steer a chosen expert set by modifying router scores at inference time, without updating model
parameters. Following \citet{fayyaz2026steermoe}, we first map raw logits to log-probabilities,
\begin{equation}
  s_t^{(\ell)} = \log\operatorname{softmax}\!\left(z_t^{(\ell)}\right),
  \label{eq:steer-logsoftmax}
\end{equation}
Let $A_+^{(m,\ell)}$ and $A_-^{(m,\ell)}$ denote disjoint expert sets to activate and deactivate
in layer $\ell$ during segment $m$. With
$s_{t,\max}^{(\ell)}=\max_j s_{t,j}^{(\ell)}$, $s_{t,\min}^{(\ell)}=\min_j s_{t,j}^{(\ell)}$, and
margin $\epsilon>0$, we set
\begin{equation}
  \widetilde s_{t,e}^{(\ell)} =
  \begin{cases}
    s_{t,\max}^{(\ell)}+\epsilon,
      & e\in A_{+}^{(m,\ell)},\\
    s_{t,\min}^{(\ell)}-\epsilon,
      & e\in A_{-}^{(m,\ell)},\\
    s_{t,e}^{(\ell)}, & \text{otherwise}.
  \end{cases}
  \label{eq:steer-margin}
\end{equation}
The model then recomputes its standard softmax, top-$k$ set, and gate normalization. Provided at most
$k$ experts are activated and at least $k$ remain eligible, the activated experts enter and the
deactivated experts leave the routed computation. Shared experts remain unchanged.
Appendix~\ref{app:steering-setup} gives the complete computation; expert sets and
intervention schedules are specified in Section~\ref{sec:steering-setup}.

\subsection{Steering Setup}
\label{sec:steering-setup}

\paragraph{Benchmark selection.}
Our benchmarks test whether behavior-targeted steering generalizes across problem difficulty and beyond
pure mathematics. The mathematics benchmarks form a difficulty progression: MATH500
\citep{hendrycks2021math} is competition-level, while AIME 2026
\citep{maa_aime} provides a more challenging competition setting. GPQA-Diamond \citep{rein2024gpqa} and
AICrypto \citep{wang2025aicrypto} then extend the evaluation to scientific and cryptographic reasoning,
respectively. For each benchmark, we choose which experts to steer, and how many, on a development
split disjoint from the test set, comparing individual experts and multi-expert combinations.
Appendix~\ref{app:steering-benchmarks} lists each development set, seed count, and token budget.

\paragraph{Model selection.}
We evaluate four open-weight MoE reasoning models: Gemma-4 \citep{gemmateam2026gemma4}, two
generations of Qwen3 \citep{team2025qwen3,team2026qwen36}, and GPT-OSS \citep{openai2025gptoss}. As summarized in Table~\ref{tab:steering-models} (Appendix~\ref{app:steering-implementation}), they differ in routed-expert pool size, routing sparsity, and the use of
shared experts. This architectural diversity tests whether behavior-targeted steering generalizes beyond
a single MoE design.

\paragraph{Evaluation setup.}
MATH500 and AIME answers are graded by DeepSeek-V4 Flash as a judge; the multiple-choice
answers of GPQA-Diamond and AICrypto are matched against the reference by rule
(Appendix~\ref{app:steering-benchmarks}).
Efficiency is measured by the number of reasoning and total generated tokens under each model's
native tokenizer.

\paragraph{Baselines.}
We compare behavior-targeted steering against three baselines:
\begin{itemize}[leftmargin=*,itemsep=2pt,topsep=3pt]
  \item \textbf{Unmodified model:} the same model with standard routing and decoding.
  \item \textbf{Random experts:} for each development-selected configuration, an equally sized random
        expert set receives the same intervention direction. For example, suppressing 200
        planning-associated experts is compared with suppressing 200 randomly selected experts.
  \item \textbf{Prompt baseline:} routing is unchanged, and a system instruction asks the model to keep
        its reasoning concise and skip unnecessary re-checks (Appendix~\ref{app:baselines}).
\end{itemize}

\begin{figure}[t]
\centering
\includegraphics[width=\textwidth]{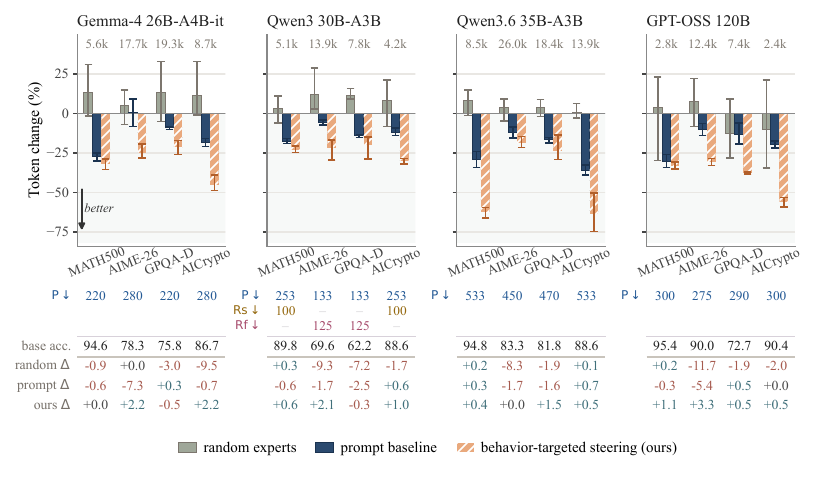}
\vspace{-10pt}
\caption{\textbf{Efficiency.} Token change relative to the unmodified model; the table reports
baseline accuracy and accuracy changes, with baseline tokens in gray. Rows above give each edit:
behavior (\textcolor{phPlanning!75!black}{P}~planning, \textcolor{phReflect!80!black}{Rf}~reflection,
and \textcolor{phRestate!75!black}{Rs}~restatement), direction,
and number of layer--expert pairs.}
\label{fig:efficiency}
\end{figure}

\subsection{Reducing Token Use while Preserving Accuracy}
\label{sec:results}
\label{sec:results-efficiency}

\paragraph{Selected expert sets shorten reasoning.}
We select each configuration on the development split and fix it for test evaluation. Steering reduces
generated tokens in all sixteen model--benchmark settings by a median of 30.4\% and by up to 63.3\%
(Qwen3.6 on AICrypto; Figure~\ref{fig:efficiency}). Accuracy is unchanged or higher in fourteen settings
and decreases by at most 0.5 percentage points in the other two. Behavior associations thus guide
useful interventions, although they do not by themselves establish what an expert causally controls.

\paragraph{Savings are largest where overthinking is.}
Models waste the most compute when they reach a solution quickly and spend thousands of tokens looking for confirmation. Steering cuts nearly twice as deep on the less demanding MATH500 and AICrypto benchmarks as it does on AIME and GPQA-Diamond (43.2\% vs.\ 24.9\% on average). On AIME, shutting down these circular re-checks even raises accuracy by 2.1 to 3.3 points in three of the four models.

\paragraph{Which experts are edited matters.}
We apply the same intervention to equally sized random expert sets. Random interventions increase
token generation in 14 of 16 settings while often degrading accuracy, so the savings come from the
selected experts rather than from perturbing the router in general.

\paragraph{Steering beats a brevity prompt.}
A natural alternative to internal interventions is simply asking the model, through a system instruction, to keep its reasoning concise
and skip unnecessary re-checks. Yet routing interventions save more tokens than this prompt in all 16 settings and match or exceed its accuracy in 14 (Appendix~\ref{app:prompt-full}).
Consistent with prior findings that reasoning models routinely ignore prompt-level length and behavioral constraints \citep{fu2026scaling,kwon2025reasonif}, natural-language instructions struggle to override the extended verification habits ingrained during post-training \citep{yueh-han2026reasoning,guo2025deepseekr1,gandhi2025cognitive}. Breaking those habits requires intervening directly on the router, yet Figure~\ref{fig:efficiency} reveals a surprising pattern: every selected expert set suppresses planning. We examine why next.

\subsection{Why Does Suppressing Planning Work?}
\label{sec:why-planning}

\begin{wrapfigure}{r}{0.47\textwidth}
\vspace{-14pt}
\centering
\includegraphics[width=0.47\textwidth]{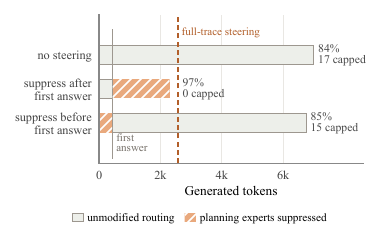}
\vspace{-16pt}
\caption{Cut at the first answer.}
\label{fig:sec54-cut}
\vspace{-9pt}
\end{wrapfigure}
At first glance, overthinking seems to be an excess of verification, so the natural target is reflection: the \emph{``Wait, let me double-check''} sentences that fill a long trace, and the behavior prior work steers away from \citep{chen2025seal}. Also, restatement looks like pure repetition, and trimming execution should shorten the derivation itself. Planning, by contrast, looks like the one behavior a model cannot do without. The
data say otherwise. On 100 MATH500 problems with Qwen3-30B-A3B, suppressing 200 planning experts cuts generated tokens by 22\%, whereas suppressing reflection or restatement experts lengthens traces (by 5\% and 1\%; the latter more than doubles the rounds of checking after the answer), and
suppressing execution experts lengthens them by 52\% or, applied more strongly, drops accuracy to 40\% (Appendix~\ref{app:steer-example}).

\paragraph{Only planning shortens the trace.}
A long trace typically reaches its answer early and then checks it again and again (Figure~\ref{fig:phase-trace-full}), each round announced
by a fresh plan: \emph{``Let me check again.''} \emph{``Alternatively, let me compute it directly.''}
\emph{``Another check:''} Suppressing each behavior acts on these rounds of checking differently:

\begin{center}
\small
\definecolor{okfill}{HTML}{FBEDE3}\definecolor{okmark}{HTML}{C9702F}\definecolor{badmark}{HTML}{800000}
\newcommand{\good}{\textcolor{okmark}{\boldmath$\checkmark$}}
\newcommand{\bad}{\textcolor{badmark}{$\times$}}
\renewcommand{\arraystretch}{1.25}
\setlength{\tabcolsep}{4pt}
\begin{tabular}{c l c >{\raggedright\arraybackslash}p{0.66\linewidth}}
\toprule
 & Suppress & Re-checks & Why\\
\midrule
\rowcolor{okfill}
\good & \nmPlanning{} & \textcolor{okmark}{fewer} & The next plan wraps up instead of opening a new round. Doubts no longer turn into
  recomputation, and the model stops after its answer.\\
\bad & \nmReflect{} & \textcolor{black!55}{same} & The model voices fewer doubts, but planning is intact. Since planning launches each
  round, the rounds still run and the model keeps verifying.\\
\bad & \nmRestate{} & \textcolor{badmark}{more} & The model can no longer reuse results from earlier steps. It re-derives them instead,
  and each re-derivation invites another check.\\
\bad & \nmExec{} & \textcolor{black!55}{same} & The model loses fluency in simple arithmetic. The rounds remain but each step slows,
  and heavy suppression can prevent an answer.\\
\bottomrule
\end{tabular}
\end{center}

\begin{tcolorbox}[
  colback=keyfill, colframe=keyfill, boxrule=0pt,
  arc=0pt, left=6pt, right=6pt, top=5pt, bottom=5pt
]
\textbf{\color{keytext}Core intuition.}
Every round of checking starts with a plan. Suppressing planning does not stop the model from planning;
it turns the next plan from opening another round into wrapping up, so whole rounds disappear together
with the behaviors inside them. 
\end{tcolorbox}

\paragraph{The change happens after the first answer.}
If planning mainly starts re-checks of an answer the model already has, suppressing it should leave the
reasoning before that answer intact and act only after it. The unmodified models reach a first answer
early, at a median of 8\% of the trace on Qwen3.6, 11\% on
Qwen3, and 21\% on Gemma-4, and 82--88\% of their planning sentences come after it. To separate the two
parts of a trace, we cut each Qwen3.6 trace on the first 100 MATH500 problems right after its first
answer and steer only one side of the cut (Figure~\ref{fig:sec54-cut}; Appendix~\ref{app:sec54}).
Before the cut, suppression leaves the reasoning almost untouched: the steered model reaches its first
answer after 422 tokens instead of 441, and what follows is as long as without steering. After the cut,
suppression alone reproduces the effect of steering the whole trace and raises accuracy from 84\% to
97\%. This shows that planning suppression primarily acts on post-answer computation rather than the reasoning that produces the first answer.

\begin{wrapfigure}{l}{0.5\textwidth}
\vspace{-12pt}
\definecolor{supfill}{HTML}{FBEDE3}\definecolor{supmark}{HTML}{C9702F}
\newcommand{\exsep}{\par\vspace{3pt}\noindent\tikz\draw[black!30, dashed, line width=0.4pt]
  (0,0)--(\linewidth,0);\par\vspace{2pt}}
\newcommand{\extag}[3]{\noindent{\setlength{\fboxsep}{1.5pt}\colorbox{#1}{\bfseries\color{#2}#3}}\par\vspace{2pt}}
\begin{tcolorbox}[colback=promptfill, colframe=black, boxrule=0.4pt, arc=0pt,
  left=6pt, right=6pt, top=5pt, bottom=5pt]
{\fontsize{7.4}{9}\selectfont\raggedright
\textbf{Problem.} Compute $99^2+99+1$ in your head.
\exsep
\extag{black!8}{black!60}{Shared prefix}
{\color{black!55}\ldots\ The answer is 9901. Let's write out the final output.
\emph{Thought process complete.} \textbf{Result:} 9901. \textbf{Step-by-step for the user:}
1.~Factor $99$ from the first two terms: $99^2+99=99(99+1)$.}
\exsep
\extag{black!12}{black!80}{No steering}
2.~Simplify inside the parentheses: $99+1=100$. 3.~Multiply: $99\times100=9900$. 4.~Add the final
term: $9900+1=9901$.\par\smallskip
\spanP{Alternative thought:} $99^2+99+1=(100-1)^2+(100-1)+1=(10000-200+1)+100-1+1=10000-200+100
\ldots$
\exsep
\extag{supfill}{supmark}{Planning suppressed}
2.~Simplify inside the parentheses: $99(100)=9900$. 3.~Add the final $1$: $9900+1=9901$.\par\smallskip
\spanP{Looks solid. I'll put the answer in the box.}\ {\color{supmark}\texttt{</think>}}\par}
\end{tcolorbox}
\vspace{-12pt}
\caption{Same prefix and seed, different next plan. More in Appendix~\ref{app:prestage-examples}.}
\label{fig:prestage}
\vspace{-8pt}
\end{wrapfigure}
\paragraph{Suppression redirects the next plan.}
The cut shows where steering acts, but not what the model does instead. In the same unmodified Qwen3.6 traces we find 314 points after the first answer where a new stage is about
to begin, rewind by one stage (about 16 tokens), and regenerate the next ${\sim}170$ tokens twice with
the same seed, once with unmodified routing and once with the planning experts suppressed. Suppression does not
silence planning; what changes is what the model plans (Figure~\ref{fig:prestage}). With unmodified routing, the first plan is
as likely to open a new angle (\emph{``Alternative thought:''}) as to wrap up (23 against 23; Table~\ref{tab:prestage}); with
the planning experts suppressed, wrap-ups outnumber new angles 68 to 11, and the model ends its thinking
(\texttt{</think>}) within these ${\sim}170$ tokens 17\% of the time instead of 5\%. Suppressing only the stage before it
still tilts the first plan toward wrapping up (43 against 21) but barely changes how often thinking
ends (6\% against 5\%): once suppression lifts, the postponed check comes back
(Appendix~\ref{app:window-example}). Therefore, steering shortens reasoning by keeping the model's next plan pointed at finishing, which is why it has to stay on after the answer.

\FloatBarrier
\section{Discussion and Conclusions}

This paper asks three questions: whether the router of a Mixture-of-Experts model reveals what a
reasoning model is doing, whether that signal generalizes across domains, and whether it can be used
to control the model's reasoning. Linear probes on router logits recover six
reasoning behaviors about as accurately as probes on the residual stream (macro-F1 0.659 vs.\ 0.658)
while using about half as many features, and the experts associated with each behavior transfer across
five domains (mean AUC 0.768). Steering these experts at inference time, without training, reduces generated tokens by a
median of 30.4\% across four models and four benchmarks, with accuracy dropping by at most 0.5 points.

The savings do not come from removing a single type of reasoning. Among the behaviors we target, only
suppressing planning-associated experts consistently shortens traces, and it does so not by preventing the model
from planning but by redirecting its next plan from opening another round of checking toward wrapping
up. MoE routing thus provides both an interpretable signal of a model's reasoning and a
training-free mechanism for steering how much computation it spends. 
Limitations are discussed in Appendix~\ref{app:limitations}. Future work can test whether the same experts govern reasoning in longer-horizon and agentic tasks.

\clearpage

\bibliography{reasoningMoE}
\bibliographystyle{iclr2027_conference}

\clearpage
\appendix

\section{Limitations}
\label{app:limitations}

\paragraph{Reliance on an LLM annotator.} Every behavior label comes from a single LLM judge (GPT-5.5)
applied with a fixed codebook, and we report no human agreement. Some boundaries are inherently
ambiguous, most often between planning and reflection when a sentence both voices a doubt and proposes
to check; a different annotator or codebook could shift these labels and the associations built on
them.

\paragraph{Efficiency, not capability.} Steering makes reasoning cheaper, but it does not make the model
reason better. Boosting the experts associated with a behavior did not reliably improve accuracy, so
the router serves here as a lever on how much a model computes rather than on whether it reaches the
right answer.

\paragraph{Model scale.} The largest model we study is GPT-OSS with 120B parameters, while open-weight
MoE models now reach several hundred billion to over a trillion parameters, with more experts per
layer. Our compute budget did not allow us to test them, so it remains uncertain whether our method is
equally effective on larger models.

\section{Additional Background}
\label{app:part-background}

Supporting material for Section~\ref{sec:data}.

\subsection{Reasoning-Behavior Taxonomy and Annotation}
\label{app:annotation}

\subsubsection{Operational Behavior Definitions}

The codebook below gives the operational boundary between each behavior and its nearest neighbors.
Labels describe the functional role of a segment, not its topic or a required position in the trace.
\label{tab:phase-codebook}

\begin{tcolorbox}[
  colback=promptfill, colframe=black, boxrule=0.4pt,
  arc=0pt, left=8pt, right=8pt, top=6pt, bottom=6pt
]
\textbf{Codebook.}\par\smallskip
{\small
\begin{description}[leftmargin=6.6em, labelwidth=6.2em, labelsep=0.4em, itemsep=2pt, topsep=0pt,
  font=\normalfont]
  \item[\nmPlanning{}] Proposes how to proceed; carrying out the proposed step is \nmExec{}.
  \item[\nmExec{}] Produces a new calculation, deduction, or construction; evaluating an existing step
    is \nmReflect{}, while importing established knowledge is \nmFact{}.
  \item[\nmRestate{}] Repeats or reformulates the problem or an established result without adding an
    inference; external knowledge is \nmFact{}.
  \item[\nmFact{}] Invokes a theorem, definition, formula, or established fact; deriving a new result
    from it is \nmExec{}.
  \item[\nmReflect{}] Checks, questions, corrects, or revises existing reasoning; ordinary forward
    progress is \nmExec{}.
  \item[\nmConclude{}] States the final answer or an intermediate result, or signals that the model is
    satisfied and done with a subgoal or the problem.
\end{description}}
\end{tcolorbox}

\subsubsection{Segmentation and Mixed-Segment Policy}

We segment traces into sentences or clauses and assign one of the six behaviors to each span.
There is no maximum-token cap: the median length is 16 tokens, with 5th/95th percentiles of 3/47 and
a maximum of 125. Character-to-token alignment drops no segments. The codebook was frozen on a
20-trace pilot before probe training; all pilot problems are restricted to the training split.

\subsubsection{Annotation Pipeline and Prompt}

\textbf{Annotation pipeline.} The pipeline has four stages: (1) sample problems and generate reasoning
traces; (2) divide each trace into sentence/clause segments; (3) ask an LLM judge (GPT-5.5, medium
reasoning effort) to label each segment with one behavior, given the segment and the full trace; and
(4) project each label onto its tokens, $y_t=y_m$ for $a_m\leq t\leq b_m$. The full trace is needed
because the same surface form can serve different functions: an equation may be newly derived,
recalled as a fact, or restated from an earlier step.
Appendix~\ref{app:trace} shows an annotated trace. During the pilot, the codebook was revised to avoid assigning
reflection from opening markers alone. Under the frozen codebook, 29\% of reflection segments lack
a doubt marker, while 13\% of execution segments contain one. Labels are not revised in response to
probe predictions.

\paragraph{Annotation prompt.} The judge receives the codebook below.

\begin{tcolorbox}[breakable,
  colback=promptfill, colframe=black, boxrule=0.4pt,
  arc=0pt, left=8pt, right=8pt, top=6pt, bottom=6pt
]
\textbf{Prompt.}\par\smallskip
{\small
\begin{description}[leftmargin=7.4em, labelwidth=7.0em, labelsep=0.4em, itemsep=3pt, topsep=0pt,
  font=\normalfont\ttfamily]
  \item[planning] Meta-statements about what to do next. Signals like ``Let's\ldots'', ``First, we
    need to\ldots'', ``Now let's try\ldots'', ``So the plan is\ldots''. Content is about the
    \emph{approach}, not the math itself.
  \item[execution] Actual derivation steps---applying a rule, doing algebra, computing a value,
    drawing a logical inference. Includes \LaTeX{} equations that are part of the derivation chain.
    NOT an opinion about a step (that is reflection).
  \item[restatement] Repeating or paraphrasing the problem statement or a previously established
    fact from earlier in the trace.
  \item[fact] Invoking external knowledge not given in the problem---a named theorem, a general
    formula, a definition.
  \item[reflection] Checks, questions, corrects, or revises existing reasoning---doubt, uncertainty,
    or a suspicion that something is wrong that triggers reconsideration, e.g.\ ``But I'm not sure
    about this.'', ``Hmm, wait.'', ``This seems off.'' If it states what to do next (``Let's try X'') $\rightarrow$ planning; if it
    computes/derives $\rightarrow$ execution; if it expresses SATISFACTION or signals wrapping up
    $\rightarrow$ conclusion.
  \item[conclusion] The final answer, an intermediate result, OR a closure/commitment signal that
    the model is satisfied and DONE reasoning---the positive ``off-ramp'', e.g.\ ``Stop. Enough.'',
    ``This confirms the earlier result.'', ``That's correct.'', ``So the answer is\ldots'',
    ``Done.''
\end{description}}
\end{tcolorbox}

\subsection{An Annotated Trace}
\label{app:trace}

Figure~\ref{fig:phase-trace-full} shows part of one annotated trace, with each segment highlighted in
the color of its behavior.

\begin{figure}[t]
\centering
\begin{tcolorbox}[
  colback=promptfill, colframe=black, boxrule=0.4pt,
  arc=0pt, left=8pt, right=8pt, top=6pt, bottom=6pt
]
\textbf{Problem.}\par\smallskip
{\itshape A man's speed with the current is 12 km/hr and the speed of the current is 2 km/hr. What is
the man's speed against the current?}

\medskip
\textbf{Trace.}\par\smallskip
{\small\baselineskip=15.5pt
\spanP{First, let me recall.}
\spanF{If a person is moving with the current, their effective speed is their own
speed in still water plus the speed of the current.}
\spanF{Conversely, when moving against the current, their effective speed is their
own speed minus the speed of the current.}
\spanV{Is that right?}
\spanP{Let me think.}
\spanE{Yeah, because when you go with the current, the current helps you, so you go faster.}
\spanE{When you go against, the current is working against you, so you go slower.}
\spanS{So, the problem gives me the speed with the current, which is 12 km/hr, and
the speed of the current is 2 km/hr.}
\spanE{So, if I let the man's speed in still water be, let's say, `v' km/hr.}
\spanE{Then, when he's going with the current, his effective speed is v + 2.}
\spanE{And that's given as 12 km/hr.}
\spanE{So, I can set up an equation: v + 2 = 12.}
\spanE{Then, solving for v, I subtract 2 from both sides: v = 12 -- 2 = 10 km/hr.}
\spanC{So, his speed in still water is 10 km/hr.}
\spanS{Now, the question is asking for his speed against the current.}
\spanE{So, that would be his speed in still water minus the speed of the current.}
\spanE{So, that would be v -- 2.}
\spanE{Since v is 10, then 10 -- 2 = 8 km/hr.}
\spanC{Therefore, his speed against the current is 8 km/hr.}
\spanC{That seems straightforward.}
\spanP{Wait, let me check if I did that right.}
\elide{57 spans omitted: the model re-derives the same answer three more ways}
\spanC{Therefore, I think 8 km/hr is the correct answer.}
\spanC{I don't see any other way to interpret it.}
\spanC{So, I think that's the answer.}
\spanC{\textbf{Final Answer} The man's speed against the current is 8 km/hr.}
\par}

\medskip
{\small \spanP{planning}\ \ \spanE{execution}\ \ \spanS{restatement}\ \ \spanF{fact}\ \
\spanV{reflection}\ \ \spanC{conclusion}}
\end{tcolorbox}
\vspace{-4pt}
\caption{\textbf{One annotated trace} (Qwen3-30B-A3B, NuminaMath problem 00245; spans 7--27 and
85--88 of 108).}
\label{fig:phase-trace-full}
\end{figure}

\section{Read: Reading Reasoning Through the Router}
\label{app:part-read}

Supporting material for Section~\ref{sec:analysis}.

\subsection{Setup}
\label{app:part1-setup}

Qwen3-30B-A3B has 48 MoE layers, 128 routed experts per layer, top-8 routing, and hidden width
2048. Problems are sampled without
replacement and split by problem, stratified by source (Table~\ref{tab:qwen-splits}).

\begin{table}[h]
\centering\small
\caption{Problem-grouped corpus splits, stratified by source.}
\label{tab:qwen-splits}
\begin{tabular}{@{}lrrr@{}}
\toprule
Source & Train & Dev & Test\\
\midrule
GSM8K training split & 100 & 22 & 28\\
NuminaMath training split & 98 & 23 & 29\\
AIME 2023 & 21 & 4 & 5\\
\midrule
Problems & 219 & 49 & 62\\
Segments & 52,008 & 10,315 & 16,632\\
\bottomrule
\end{tabular}
\end{table}

\paragraph{Probes.}
Router logits are the continuous pre-top-$k$ scores, centered within each token; residual features are
the block-input hidden states at the same positions. Every probe is an L2-regularized multinomial
logistic regression with training-fit standardization. The regularization strength
$C\in\{0.0003,\ldots,30\}$ and the layer are selected on dev macro-F1 for the corresponding evaluation
unit, and the test split is used only to report frozen configurations. Token probes use up to 200
labeled tokens per trace (43,800 / 9,800 / 12,400 train / dev / test tokens); segment probes mean-pool
each representation within segment boundaries and use all segments. Confidence intervals come from
2,000 problem-bootstrap resamples.

\subsection{Additional Results}
\label{app:part1-results}

\paragraph{Layer window and width-matched control.}
\label{app:router-depth}
The router window concatenates centered logits from each layer as a separate 128-dimensional block. Its
center, L12, is selected on dev and its radius of four layers is fixed in advance. Widening the window
changes little (Table~\ref{tab:router-depth}). To check that parity is not due to drawing on nine
layers, we also concatenate residual states from the same L8--L16, each projected to 128 dimensions
with a train-fit PCA. At equal width and equal layer coverage, router and residual token F1 are 0.659 and 0.657 (paired
difference 0.2 points, 95\% CI [$-0.6$, 1.0]); at the segment level they are 0.731 and 0.736
($-0.5$ points, [$-1.4$, 0.2]).

\begin{table}[h]
\centering\small
\caption{Router layer windows centered on L12 (macro-F1).}
\label{tab:router-depth}
\begin{tabular}{@{}lrrr@{}}
\toprule
Layers & Width & Token test & Segment test\\
\midrule
L12 only & 128 & 0.608 & 0.708\\
L8--L16 ($\pm4$) & 1,152 & 0.659 & 0.731\\
L7--L17 ($\pm5$) & 1,408 & 0.656 & 0.731\\
L6--L18 ($\pm6$) & 1,664 & 0.655 & 0.733\\
L5--L19 ($\pm7$) & 1,920 & 0.653 & 0.732\\
\bottomrule
\end{tabular}
\end{table}

\paragraph{Pooling within a segment.}
Table~\ref{tab:qwen-readout} compares ways of turning router states into a segment prediction at L12,
without neighboring segments. Pooling features before classification outperforms aggregating
token-level predictions. Mean+max pooling adds 0.9 F1 points over mean pooling (paired 95\% CI
[0.4, 1.4]) at twice the feature width, so the main text uses the compact mean-pooled readout.

\begin{table}[h]
\caption{Segment readout from router L12.}
\label{tab:qwen-readout}
\centering\small
\setlength{\tabcolsep}{4pt}
\begin{tabular}{@{}lcc@{}}
\toprule
Readout & Width & Macro-F1\\
\midrule
Token majority vote & 128 & 0.608\\
Token probability average & 128 & 0.613\\
Final token & 128 & 0.702\\
Max pooling & 128 & 0.682\\
Mean pooling & 128 & 0.708\\
Mean + max pooling & 256 & 0.716\\
\bottomrule
\end{tabular}
\end{table}

\section{Identify: Identifying Reasoning Experts}
\label{app:part-identify}

Supporting material for Section~\ref{sec:identify}.

\subsection{Association Scores and Transfer Protocol}
\label{app:expert-localization}

\paragraph{Corpus.} The association analysis uses 330 Qwen3-30B-A3B traces from each of six sources
(Table~\ref{tab:qwen-domains}). Expert selections are recorded during generation.

\begin{table}[h]
\centering\small
\caption{Cross-dataset association corpus, with 330 problems per source.}
\label{tab:qwen-domains}
\begin{tabular}{@{}llr@{}}
\toprule
Source & Domain & Segments\\
\midrule
Mathematical mixture & Mathematics & 78,955\\
MATH500 \citep{hendrycks2021math} & Mathematics & 79,816\\
LiveCodeBench \citep{jain2024livecodebench} & Code & 187,068\\
GPQA (non-Diamond) \citep{rein2024gpqa} & Science & 112,825\\
MMLU-Pro \citep{wang2024mmlupro} & Knowledge & 53,770\\
CommonsenseQA \citep{talmor2019commonsenseqa} & Commonsense & 17,123\\
\bottomrule
\end{tabular}
\end{table}

\paragraph{Association.} Expert profiles use the dataset-relative risk differences of
Equation~\ref{eq:expert-specificity}; top-10 selection and profile comparisons are made within each
layer. The permutation null shuffles labels within traces, preserving trace membership and label counts.
The descriptive layer, L12, maximizes the same- minus different-behavior correlation gap; L24 and L36
are nearly tied, so L12 is a best-observed summary rather than a validated choice.

\paragraph{Transfer.} For each source--target pair and behavior, the source-selected top-10 set scores
target segments by its selection frequency, giving 180 directed comparisons per layer (six sources,
five targets, six behaviors). The supervised transfer probe concatenates segment-mean selection
fractions from layers 3, 11, 23, and 36 (512 dimensions); its cross-domain macro-F1 of 0.561 compares
with 0.701 within a dataset and 0.096 for always predicting the most frequent behavior.

\section{Steer: Steering Experiments}
\label{app:part-steer}

Supporting material for Section~\ref{sec:intervention}.

\subsection{Steering Protocol and Reproducibility}
\label{app:steering-setup}

\subsubsection{Activation and Deactivation Implementation}
\label{app:steering-implementation}

Table~\ref{tab:steering-models} lists the routing configuration of each model; the top-$k$ value
and the presence of shared experts determine how many experts an intervention can activate.

\begin{table}[!h]
\centering
\caption{MoE models and routing configurations used for steering.}
\label{tab:steering-models}
\small
\setlength{\tabcolsep}{3pt}
\begin{tabular}{@{}lccccc@{}}
\toprule
Model & Layers & Total & Active & Routed & Top-$k$ \& shared\\
\midrule
Gemma-4-26B-A4B-it & 30 & 26B  & 4B         & 128 & 8\\
Qwen3-30B-A3B      & 48 & 30B  & 3B         & 128 & 8\\
Qwen3.6-35B-A3B    & 40 & 35B  & 3B         & 256 & 8 $+$ 1 shared\\
GPT-OSS-120B       & 36 & 120B & $\sim$5.1B & 128 & 4\\
\bottomrule
\end{tabular}
\end{table}

After applying Equation~\ref{eq:steer-margin}, we recompute the routing distribution and selected set,
\begin{equation}
  \widetilde p_t^{(\ell)}
    =\operatorname{softmax}\!\left(\widetilde s_t^{(\ell)}\right),
  \qquad
  \widetilde S_t^{(\ell)}
    =\operatorname{TopK}\!\left(\widetilde p_t^{(\ell)},k\right).
  \label{eq:steer-reroute}
\end{equation}
The selected probabilities are then renormalized and passed through the standard routed experts:
\begin{equation}
  \overline g_{t,e}^{(\ell)}=
    \frac{\widetilde p_{t,e}^{(\ell)}}
         {\sum_{j\in\widetilde S_t^{(\ell)}}\widetilde p_{t,j}^{(\ell)}},
  \qquad
  \widetilde{\operatorname{MoE}}^{(\ell)}(h_t^{(\ell)})
    =\sum_{e\in\widetilde S_t^{(\ell)}}
      \overline g_{t,e}^{(\ell)}E_e^{(\ell)}(h_t^{(\ell)}).
  \label{eq:steer-output}
\end{equation}
At most $k$ experts are activated in any layer. If a behavior yields more candidates, we retain those
with the highest association scores. Deactivation removes experts from the routed path; shared experts
remain unchanged.

\subsubsection{Benchmarks, Prompts, and Scoring}
\label{app:steering-benchmarks}

All models and conditions share the same decoding settings. For each benchmark, the list below gives
the development set on which the intervention is selected (disjoint from the test benchmark), the
number of seeds, and the maximum new tokens for thinking and for the answer.
\begin{itemize}[leftmargin=*,itemsep=2pt,topsep=3pt]
  \item \textbf{MATH500:} development set from the MATH training split; 3 seeds; \mbox{16,384 / 4,096} tokens.
  \item \textbf{AIME 2026:} development set of 5 AIME 2025 and 10 HMMT 2026 problems; 8 seeds; \mbox{28,672 / 4,096} tokens.
  \item \textbf{GPQA-Diamond:} development set from MMLU-Pro \citep{wang2024mmlupro}; 3 seeds;
        \mbox{24,576 / 4,096} tokens.
  \item \textbf{AICrypto:} the MATH500 development set; 3 seeds; \mbox{16,384 / 4,096} tokens.
\end{itemize}

\paragraph{Answer grading.} For MATH500 and AIME 2026, the judge compares the extracted final answer
with the reference using the prompt below.

\begin{tcolorbox}[breakable,
  colback=promptfill, colframe=black, boxrule=0.4pt,
  arc=0pt, left=8pt, right=8pt, top=6pt, bottom=6pt
]
\textbf{Prompt.}\par\smallskip
You are an expert grader for competition mathematics. Your task is to decide whether a candidate final
answer is mathematically equivalent to the reference answer.\par\smallskip
Reference answer: \texttt{\{gold\}}\\
Candidate answer: \texttt{\{ans\}}\par\smallskip
Grading criteria:
\begin{enumerate}[leftmargin=1.6em, itemsep=1pt, topsep=2pt]
  \item Judge mathematical equivalence, not surface form: two answers are equivalent if and only if they
        denote the same value or mathematical object. Accept any correct notation, including equivalent
        fractions, radicals, decimals that are exactly equal in value, factored or expanded expressions,
        and differences in \LaTeX{} formatting, spacing, or simplification.
  \item Do NOT accept numerical approximations of an exact reference answer (e.g., 0.707 for
        sqrt(2)/2), unless the reference answer is itself given as that approximation.
  \item For answers consisting of multiple components (sets, tuples, lists), all components must match;
        ignore ordering only where the mathematical object is unordered.
  \item If the candidate answer is ambiguous, incomplete, or contains multiple conflicting final
        answers, grade it as not equivalent.
\end{enumerate}
Respond with exactly one word: YES or NO.
\end{tcolorbox}

GPQA-Diamond and AICrypto are multiple choice and need no judge: the options follow the question, and
a regular expression extracts the choice the model puts in \verb|\boxed{}| and compares it with the
reference. GPQA-Diamond has one correct letter among four; AICrypto options are 0-indexed, include
``None of the above'', and may have several correct answers, all of which must be selected. The
appended instructions are:

\begin{tcolorbox}[
  colback=promptfill, colframe=black, boxrule=0.4pt,
  arc=0pt, left=8pt, right=8pt, top=6pt, bottom=6pt
]
\textbf{GPQA-Diamond.}\par\smallskip
Put your final answer as the option letter (A, B, C, or D) in \verb|\boxed{}|.\par\medskip
\textbf{AICrypto.}\par\smallskip
Some questions may have multiple correct answers; select all that apply. Provide the 0-indexed
integer(s) of the correct option(s), comma-separated. Put your final answer in \verb|\boxed{}|
(e.g.\ \verb|\boxed{0}| or \verb|\boxed{0,2,3}|).
\end{tcolorbox}

\subsubsection{Baselines, Controls, and Efficiency Measurement}
\label{app:baselines}

\paragraph{Prompt baseline.}
The prompt baseline asks in natural language for concise reasoning: move on once a step is
established, avoid unnecessary re-checks, and commit once confident. Decoding settings are unchanged;
only the system prompt below is added.

\begin{tcolorbox}[
  colback=promptfill, colframe=black, boxrule=0.4pt,
  arc=0pt, left=8pt, right=8pt, top=6pt, bottom=6pt
]
\textbf{Prompt.}\par\smallskip
Solve the problem step by step, but keep the reasoning concise and efficient. Once a step is well
established, move on rather than performing unnecessary repeated checks or re-verifying points that are
already sufficiently supported. Once you are confident the answer is correct, commit to it.\par
Accuracy comes first: concision should never come at the expense of completeness.
\end{tcolorbox}

\subsection{Additional Steering Results}
\label{app:steering-results}

\subsubsection{Full Accuracy--Efficiency Results}
\label{app:prompt-full}

Table~\ref{tab:prompt-full} lists the token and accuracy changes of the prompt baseline and of
behavior-targeted steering for every model and benchmark, the values plotted in
Figure~\ref{fig:efficiency}.

\begin{table}[t]
\centering\small
\caption{\textbf{Prompt baseline versus behavior-targeted steering} for all models. Changes are relative
to the unmodified model; token changes average the evaluated seeds, as in Figure~\ref{fig:efficiency}.}
\label{tab:prompt-full}
\begin{tabular}{@{}ll rr rr@{}}
\toprule
 & & \multicolumn{2}{c}{$\Delta$ tokens (\%)} & \multicolumn{2}{c}{$\Delta$ accuracy (pp)}\\
\cmidrule(lr){3-4}\cmidrule(l){5-6}
Model & Benchmark & Prompt & Ours & Prompt & Ours\\
\midrule
Gemma-4 & \texttt{MATH500} & $-27.3$ & $-32.2$ & $-0.6$ & $+0.0$\\
 & \texttt{AIME} & $+0.6$ & $-24.9$ & $-7.3$ & $+2.2$\\
 & \texttt{GPQA-D} & $-9.1$ & $-21.2$ & $+0.3$ & $-0.5$\\
 & \texttt{AICrypto} & $-18.6$ & $-44.9$ & $-0.7$ & $+2.2$\\
\midrule
Qwen3 & \texttt{MATH500} & $-17.4$ & $-23.1$ & $-0.6$ & $+0.6$\\
 & \texttt{AIME} & $-5.7$ & $-22.0$ & $-1.7$ & $+2.1$\\
 & \texttt{GPQA-D} & $-14.0$ & $-19.8$ & $-2.5$ & $-0.3$\\
 & \texttt{AICrypto} & $-12.1$ & $-30.0$ & $+0.6$ & $+1.0$\\
\midrule
Qwen3.6 & \texttt{MATH500} & $-29.3$ & $-62.4$ & $+0.3$ & $+0.4$\\
 & \texttt{AIME} & $-12.0$ & $-18.6$ & $-1.7$ & $+0.0$\\
 & \texttt{GPQA-D} & $-16.7$ & $-23.9$ & $-1.6$ & $+1.5$\\
 & \texttt{AICrypto} & $-36.0$ & $-63.3$ & $+0.7$ & $+0.5$\\
\midrule
GPT-OSS & \texttt{MATH500} & $-30.3$ & $-33.3$ & $-0.3$ & $+1.1$\\
 & \texttt{AIME} & $-10.0$ & $-30.9$ & $-5.4$ & $+3.3$\\
 & \texttt{GPQA-D} & $-13.2$ & $-37.6$ & $+0.5$ & $+0.5$\\
 & \texttt{AICrypto} & $-19.5$ & $-56.0$ & $+0.0$ & $+0.5$\\
\bottomrule
\end{tabular}
\end{table}

\subsubsection{One Problem Under Each Suppression}
\label{app:steer-example}

We follow MATH500 problem 29 on Qwen3-30B-A3B, suppressing one behavior's experts at a time
(Table~\ref{tab:steer-example}). The problem asks how many
battalions of 4 upper-class and 8 lower-class soldiers can be drawn from 5 and 10; the answer is
$\tbinom54\tbinom{10}8=5\cdot45=225$. The excerpts below start at each trace's answer. Bracketed numbers
index sentences and [\,\ldots\,] marks omitted text; \spanP{planning} sentences start a round of
checking and \spanV{reflection} sentences voice a doubt. Each row is a single sample, but its direction
matches the 100-problem means in Section~\ref{sec:why-planning}.

\begin{table}[H]
\centering\small
\caption{\textbf{One problem under each suppression.} Parentheses give the number of suppressed
layer--expert pairs. Re-checks: rounds of checking after the answer; Doubts and Openers: reflection
and planning sentences in the trace.}
\label{tab:steer-example}
\renewcommand{\arraystretch}{1.12}
\begin{tabular}{@{}l r c c c c c@{}}
\toprule
Suppress & Tokens & Answer at & Re-checks & Doubts & Openers & Correct\\
\midrule
none & 1,494 & [31] & 4 & 7 & 10 & yes\\
\rowcolor{keyfill}
planning ($\downarrow$200) & 1,076 & [30] & 2 & 3 & 7 & yes\\
reflection ($\downarrow$100) & 1,579 & [38] & 4 & 3 & 15 & yes\\
restatement ($\downarrow$100) & 1,866 & [43] & 5 & 7 & 15 & yes\\
execution ($\downarrow$30) & 1,758 & [64] & 4 & 5 & 14 & yes\\
execution ($\downarrow$100) & 16,384 & -- & -- & 19 & 40 & no\\
\bottomrule
\end{tabular}
\end{table}

\begin{tcolorbox}[breakable,
  colback=promptfill, colframe=black, boxrule=0.4pt,
  arc=0pt, left=8pt, right=8pt, top=6pt, bottom=6pt
]
{\small
\textbf{No suppression.} Every doubt launches a recomputation.\par\smallskip
[31] Therefore, the total number of different battalions is 5 * 45 = 225.
\spanV{[32] Wait, is that all?} \spanP{[33] Let me check again.} [34] So upper class: 5 choose 4 is 5.
[35] Lower class: 10 choose 8 is 45. [\,\ldots\,]
\spanP{[38] But let me verify the combination calculations once more to be sure.} [\,\ldots\,]
[44] Alternatively, as I did before, 10 choose 2. [45] Which is 45.
\spanP{[46] Alternatively, if I compute C(10, 8) directly:} [47--50] C(10, 8) = 10! / (8! * 2!) = 90 / 2 = 45.
[\,\ldots\,] \spanP{[59] Another check: total number of ways to choose 4 upper and 8 lower.} [\,\ldots\,]
[67] Therefore, I think the answer is 225.

\medskip
\textbf{Planning suppressed ($\downarrow$200): fewer re-checks.} Doubts are still voiced, but each is settled in a sentence,
and earlier results are quoted rather than re-derived.\par\smallskip
[30] So the answer should be 225 different battalions.
\spanV{[31] Wait, let me check if there's any mistake here.}
\spanV{[32] For example, is there any restriction on the soldiers?}
[33] The problem says it's just about selecting 4 upper from 5 and 8 lower from 10. [\,\ldots\,]
\spanV{[36] Another thought: sometimes problems might have hidden constraints,} [\,\ldots\,]
[37] So, 5 choose 4 is 5, 10 choose 8 is 45, 5*45=225. [38] Yeah, that seems right. [\,\ldots\,]
[43] Therefore, I think 225 is the correct answer.

\medskip
\textbf{Reflection suppressed ($\downarrow$100): same re-checks.} Doubts fall from 7 to 3, but planning sentences rise from
10 to 15 and launch the same rounds; 45 is derived twice more.\par\smallskip
[38] So, 5 $\times$ 45 = 225. \spanV{[39] Wait, is that all?}
\spanP{[40] Let me check if there's any other factor I need to consider.} [\,\ldots\,]
\spanP{[45] Another thought: sometimes, people make mistakes with combinations.} [\,\ldots\,]
[49--51] C(10, 2) = 10! / (2! * 8!) = (10 $\times$ 9)/2 = 45. [\,\ldots\,]
\spanP{[59] But let me check again.} [\,\ldots\,] [67] So, the number of ways is C(10, 2) = 45.
\spanP{[71] But wait, maybe I need to verify once again.} [\,\ldots\,] [76] Therefore, I think the answer is 225.

\medskip
\textbf{Restatement suppressed ($\downarrow$100): more re-checks.} The model cannot close a check by quoting an earlier
result, so it re-derives it, and each re-derivation invites another doubt.\par\smallskip
[43] Therefore, the number of different battalions is 225.
\spanV{[44] Wait, but let me make sure I didn't make any miscalculations.}
\spanP{[45] Let me verify the combination numbers again.} [\,\ldots\,] [52] So, 10*9/2 = 45.
\spanP{[55] Is there another way to think about this?} [\,\ldots\,]
\spanP{[61] Another check: suppose there were 5 upper and 10 lower.} [62] If we needed to choose 4 upper
and 8 lower. [63] So for upper, 5C4 is 5, and for lower, 10C8 is 45. [64] So total is 5*45=225.
\spanP{[65] Alternatively, if I think about the lower class soldiers: 10 soldiers, choosing 8.} [\,\ldots\,]
[68] So 10C2 is 45. [\,\ldots\,]
\spanV{[79] Wait, another thought: are there any restrictions on the number of soldiers?} [\,\ldots\,]
[85] So, I think the answer is 225.

\medskip
\textbf{Execution suppressed ($\downarrow$30): same re-checks, slower steps.} The plan is unchanged but arithmetic
slows, and the answer arrives at [64] instead of [31]:\par\smallskip
[37] 10! [38] is 3,628,800. [39] 8! [40] is 40320, and 2! [41] is 2. [42] So 3,628,800 divided by
(40320 * 2) = 3,628,800 / 80,640. [\,\ldots\,] [47] 80,640 * 40 is 3,225,600. [48] 80,640 * 5 is 403,200.
[49] So total is 3,225,600 + 403,200 = 3,628,800. [50] Yes!\par\smallskip
Suppressing 100 execution pairs instead, the model never gets past $\tbinom54$ and ends at the token limit repeating
``96,480 * 36 = 3, 96,480 * 36 = 3, \ldots'' without an answer.
}
\end{tcolorbox}

\subsubsection{The Next Plan: Counts and Examples}
\label{app:prestage-examples}

Table~\ref{tab:prestage} summarizes the experiment of Figure~\ref{fig:prestage} over all 314 cut
points. The first planning sentence of each continuation is classified by keyword as a wrap-up (box,
final, format, \ldots) or a new angle (alternative, another way, interpretation, \ldots); cut points
whose first plan is neither are not counted. The window condition suppresses planning only for the
preceding stage or 100 tokens and then releases it.

\begin{table}[H]
\centering\small
\caption{\textbf{First plan after the cut} (Qwen3.6, MATH500, 314 cut points). The last column is the
share of continuations that end thinking within ${\sim}170$ tokens.}
\label{tab:prestage}
\begin{tabular}{@{}lccc@{}}
\toprule
Planning experts & Wrap-up & New angle & Thinking ends\\
\midrule
unmodified & 23 & 23 & 5\%\\
suppressed throughout & 68 & 11 & 17\%\\
suppressed for one window & 43 & 21 & 6\%\\
\bottomrule
\end{tabular}
\end{table}

Two more cut points, with the same prefix and seed in both conditions; highlighted is the first
planning sentence.

\begin{tcbraster}[raster columns=2, raster equal height, raster column skip=6pt,
  colback=promptfill, colframe=black, boxrule=0.4pt, arc=0pt,
  left=6pt, right=6pt, top=5pt, bottom=5pt]
\begin{tcolorbox}
{\small\raggedright
\textbf{Find $(a,b)$ with $f(x)-f(a)=(x-a)^2(x-b)$.}\par\smallskip
{\color{black!55}\ldots\ So $a=-2$ and $b=1$ are correct. Let's double check the question for any
other constraints.}\par\smallskip
\textbf{No steering}\par \spanP{Alternative approach:} $x=a$ is a root of multiplicity at least 2, so
the derivative must also vanish at $x=a$ \ldots\par\smallskip
\textbf{Planning suppressed}\par Everything is consistent. Final Answer: $\boxed{(-2,1)}$.\
\texttt{</think>}\par}
\end{tcolorbox}
\begin{tcolorbox}
{\small\raggedright
\textbf{Find the roots of $(x-3)^3+(x-7)^3=(2x-10)^3$.}\par\smallskip
{\color{black!55}\ldots\ $(x-3)(x-5)(x-7) = x^3 - 15x^2 + 71x - 105$.}\par\smallskip
\textbf{No steering}\par Yes, it matches perfectly. \spanP{Refining the substitution explanation:} Let
$u = x-3$ and $v = x-7$ \ldots\par\smallskip
\textbf{Planning suppressed}\par Matches perfectly! Conclusion: the roots are $x = 3, 5, 7$.
$\boxed{3,5,7}$\ \texttt{</think>}\par}
\end{tcolorbox}
\end{tcbraster}

\subsubsection{Suppressing Only a Window}
\label{app:window-example}

Suppressing planning for a short window (the preceding stage, or 100 tokens) tilts the next plan
toward wrapping up, but the postponed check returns once suppression lifts
(Section~\ref{sec:why-planning}). Qwen3.6 on MATH500 problem 78, same prefix and seed, with the planning
experts suppressed for 100 tokens after the verified answer $(a,b)=(-2,1)$ and then released:

\begin{tcolorbox}[breakable,
  colback=promptfill, colframe=black, boxrule=0.4pt,
  arc=0pt, left=8pt, right=8pt, top=6pt, bottom=6pt
]
{\small
{\color{black!55}Shared prefix: \ldots\ So the values $a=-2$ and $b=1$ are correct. The ordered pair is
$(a,b)=(-2,1)$. Let's}\par\medskip
\textbf{No steering.} \spanP{double check if there could be other interpretations.} This implies that
$x=a$ is a root of multiplicity at least 2 for $f(x)-f(a)$. [\,\ldots\,] For $x=a$ to be a root of
multiplicity 2, the derivative $f'(x)$ must also be zero at $x=a$. [\,\ldots\,]
$3a(a+2)=0$. So $a=0$ or $a=-2$.\par\medskip
\textbf{Planning suppressed, first 100 tokens.} check the question: ``Enter the ordered pair $(a,b)$.''
The answer should be $(-2,1)$. Is there any other constraint? [\,\ldots\,] So the answer is indeed
$(-2,1)$. Done. \spanP{I will put \texttt{\textbackslash boxed\{(-2, 1)\}} in the final answer.}\par\medskip
\textbf{After release.} \spanP{Double checking the steps just to be absolutely sure.} [\,\ldots\,]
Note that $f(x)-f(a)$ has a root at $x=a$. \spanP{Let's check the multiplicity of the root} $x=a$ in
$f(x)-f(a)$. [\,\ldots\,]
}
\end{tcolorbox}

While suppressed, the model checks the constraints and commits to the answer; as soon as suppression
lifts, it opens the same double-root check the unmodified run started 100 tokens earlier.

\subsubsection{Section~\ref{sec:why-planning} Details}
\label{app:sec54}

\paragraph{Setup.}
Each model answers the first 100 MATH500 problems once, with the sampling settings of
Appendix~\ref{app:steering-benchmarks} and a 16,384-token limit. The steered condition applies the
MATH500 expert set of Figure~\ref{fig:efficiency} at every generated token.
Sentences are labeled with the annotator and codebook of Section~\ref{sec:data}, and answers are graded
as in Appendix~\ref{app:steering-benchmarks}.

\paragraph{Cut at the first answer.}
For Qwen3.6, each unmodified trace is cut right after its first answer (median position 8\% of the
trace) and continued either without steering or with the planning experts suppressed. A third arm
starts from the trace generated under suppression up to its first answer and continues without
steering; two traces without a conclusion sentence are skipped there.

\end{document}